\documentclass[letterpaper, 10pt, conference]{IEEEtran}

\IEEEoverridecommandlockouts

\pdfoutput=1

\usepackage{cite}
\usepackage{amsmath,amssymb,amsfonts}
\usepackage{graphicx}
\usepackage{textcomp}
\usepackage{xcolor}

\usepackage{algpseudocode}
\usepackage{subcaption}
\usepackage{booktabs}

\def\BibTeX{{\rm B\kern-.05em{\sc i\kern-.025em b}\kern-.08em
    T\kern-.1667em\lower.7ex\hbox{E}\kern-.125emX}}

\begin{document}

\title{Enhancing Multi-Drone Coordination for Filming Group Behaviours in Dynamic Environments\\

}

\author{%
Aditya Rauniyar$^{1}$, Jiaoyang Li$^{1}$, and Sebastian Scherer$^{1}$.

\thanks{$^{1}$ A. Rauniyar, J. Li, and S. Scherer are with the Robotics Institute, School of Computer Science at Carnegie Mellon University, Pittsburgh, PA, USA
        {\tt\small \{arauniya, jiaoyanl, basti\}@andrew.cmu.edu}}%
\thanks{This work is supported by the National Science Foundation under Grant No. 2024173.}%
}


\maketitle

\begin{abstract}
Multi-Agent Path Finding (MAPF) is a fundamental problem in robotics and AI, with numerous applications in real-world scenarios. One such scenario is filming scenes with multiple actors, where the goal is to capture the scene from multiple angles simultaneously. Here, we present a formation-based filming directive of task assignment followed by a Conflict-Based MAPF algorithm for efficient path planning of multiple agents to achieve filming objectives while avoiding collisions. We propose an extension to the standard MAPF formulation to accommodate actor-specific requirements and constraints. Our approach incorporates Conflict-Based Search, a widely used heuristic search technique for solving MAPF problems. We demonstrate the effectiveness of our approach through experiments on various MAPF scenarios in a simulated environment. The proposed algorithm enables the efficient online task assignment of formation-based filming to capture dynamic scenes, making it suitable for various filming and coverage applications.

\end{abstract}


\section{Introduction}
Videography in itself has matured quite a built in recent times to bring froth the immersing experience to its audience. These include high-skill tasks and come with experience on the field, if someone was to do this by themselves, performing a task and recording at the same time, this would be immensely challenging. Such situation especially today are more prone as the content creators' community is rising in platforms like youtube. This leads to much-needed progress on dynamic capture with filming effects of a subject performing. Current breakthroughs in autonomous unmanned aerial vehicles (UAVs) now has given rise to using such systems for aerial cinematography. Hence, in order to capture the best shots, multi-view filming has been proven to best capture and present to its audience as seen in most of the movies today. 

Filming dynamic scenes with multiple actors from diverse viewpoints is a challenging task that requires careful planning and coordination. Traditional filming techniques often involve fixed camera positions, limiting the capturing of complex group behaviors and interactions. Unmanned aerial vehicles (UAVs) offer a promising solution by providing the flexibility to capture dynamic scenes from various angles. In this report, we present a comprehensive approach for capturing group behaviors in filming scenarios by dynamically assigning viewpoints to UAVs using Multi-Agent Path Finding (MAPF) techniques.

\begin{figure*}
\centering
    \includegraphics[scale = 0.5]{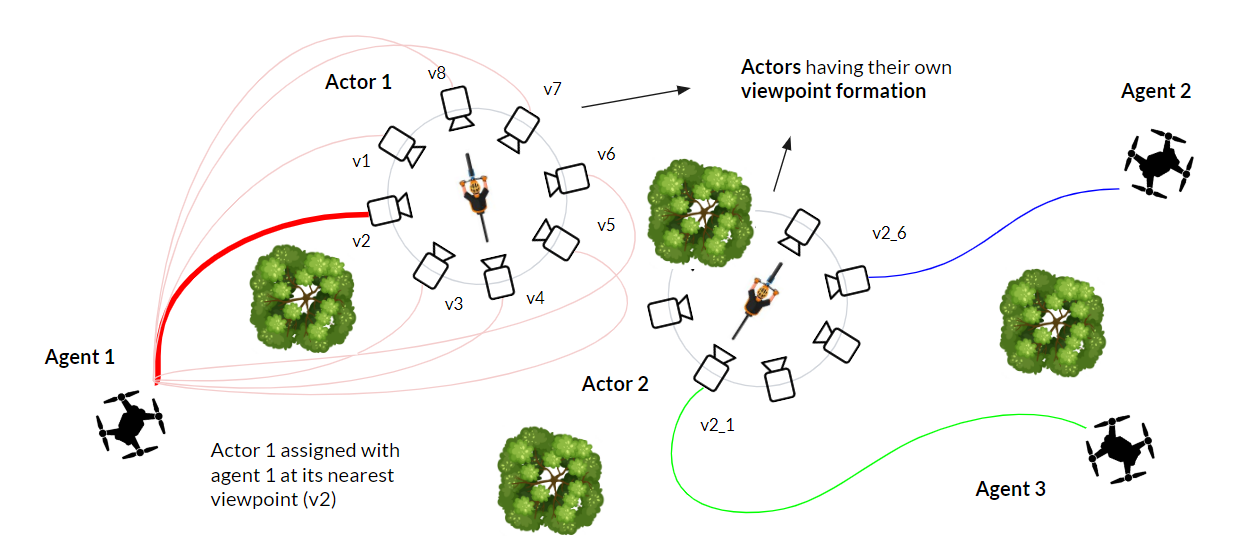}
\caption{The image showcases a scenario with actors (A) and agents/drones (D) in the scene. The agents' trajectories are designed to approach the nearest diverse viewpoint (V) from the set of possible viewpoints. This notation highlights the actors, agents, and viewpoint elements involved in the tracking process.}
\label{fig:Task_Assignment}
\end{figure*}

The primary objective of our approach is to create diverse viewpoints that effectively capture the group behaviors of actors in a scene. We address two key challenges: task creation for selecting optimal viewing angles for different actors, and task assignment to assign UAVs with appropriate viewpoints. These challenges involve considering factors such as scene composition, actor behavior, desired cinematographic effects, and the capabilities of the UAVs.

In the task creation phase, we focus on determining the optimal viewing angles for each actor. This process involves careful consideration of various factors to capture the essence of the group behaviors. By selecting a set of diverse viewpoints, we ensure that the resulting footage provides a comprehensive and engaging representation of the scene. This task creation phase is crucial in laying the foundation for capturing dynamic group behaviors effectively.

Once the optimal viewing angles are determined, the task assignment phase comes into play. This phase involves assigning UAVs to the selected viewpoints based on their capabilities and availability. The goal is to achieve efficient and coordinated filming while ensuring that each UAV captures its assigned viewpoint successfully. This task assignment process requires addressing challenges such as UAV trajectory planning, collision avoidance, and synchronization with actor movements.

To address these challenges, we leverage Conflict-Based Search (CBS), an efficient algorithm for MAPF, and extend it to incorporate actor-specific requirements and constraints. The CBS MAPF algorithm facilitates the coordination of UAV movements and actor behaviors, ensuring collision-free paths and satisfying the viewpoint assignment constraints. By integrating CBS MAPF with dynamic viewpoint assignment, we enable the efficient capture of group behaviors in dynamic filming scenarios.

Through extensive simulations and evaluations on various filming scenarios, we demonstrate the effectiveness of our approach in capturing diverse group behaviors with dynamic viewpoints. Our approach not only enhances the visual quality of the captured footage but also enables filmmakers to convey complex interactions and behaviors within a group effectively.

The rest of this report is organized as follows: In Section 2, we provide a detailed overview of related work in the field of dynamic viewpoint assignment and MAPF for filming applications. Section 3 presents the methodology and approach used in our proposed solution, including the task creation and task assignment phases. In Section 4, we describe the CBS MAPF algorithm and its integration with dynamic viewpoint assignment. Section 5 presents the experimental setup and evaluation results, showcasing the effectiveness of our approach through simulations. Finally, in Section 6, we conclude the report and discuss future research directions.

Overall, our proposed approach offers a novel and efficient solution for capturing group behaviors in filming scenarios. By dynamically assigning viewpoints to UAVs and leveraging CBS MAPF, we enable filmmakers to capture diverse and captivating visual perspectives of dynamic scenes with multiple actors.

\section{Related Work}
The field of autonomous aerial cinematography has witnessed significant advancements in recent years. Bonatti et al. \cite{1} introduced the concept of autonomous aerial cinematography in unstructured environments, incorporating learned artistic decision-making for capturing visually appealing shots. Bucker et al. \cite{2} explored the coordination of multiple aerial cameras for robot cinematography, enabling synchronized and collaborative filming.

To address the challenges of capturing group behaviors in filming scenarios, Bonatti et al. \cite{3} proposed a robust aerial cinematography platform that focuses on localizing and tracking moving targets in unstructured environments. Ho et al. \cite{4} presented a method for 3D human reconstruction in the wild using collaborative aerial cameras, allowing for comprehensive capture of group behaviors.

The autonomous drone cinematographer framework introduced by Bonatti et al. \cite{5} leveraged artistic principles to generate smooth, safe, and occlusion-free trajectories for aerial filming, enhancing the visual quality and capturing engaging shots.

In the context of multi-agent pathfinding (MAPF), Kottinger et al. \cite{6} proposed a conflict-based search algorithm for multi-robot motion planning with kinodynamic constraints. This algorithm is leveraged in our approach to coordinate UAV movements and actor behaviors effectively. Sharon et al. [8] introduced conflict-based search as an efficient algorithm for optimal multi-agent pathfinding, which serves as the foundation for our task assignment phase.

In the domain of multi-drone multi-target tracking, Liu et al. [7] presented a benchmark for robust multi-drone tracking to resolve target occlusion, contributing to the challenges of capturing dynamic scenes with multiple actors.

Krátký et al. [9] explored the concept of autonomous aerial filming with distributed lighting by a team of unmanned aerial vehicles, showcasing the potential for synchronized and coordinated filming using UAVs.

In this report, we propose a comprehensive approach for capturing group behaviors in filming scenarios by dynamically assigning viewpoints to UAVs using Multi-Agent Path Finding (MAPF) techniques. Our approach builds upon the concepts and techniques introduced by the aforementioned studies, addressing the challenges of dynamic viewpoint assignment, task creation, and efficient coordination of UAV movements.

The remainder of this report is organized as follows: In Section 2, we provide a detailed overview of related work in the field of dynamic viewpoint assignment and MAPF for filming applications, citing the relevant studies. Section 3 presents the methodology and approach used in our proposed solution, including the task creation and task assignment phases. In Section 4, we describe the CBS MAPF algorithm and its integration with dynamic viewpoint assignment. Section 5 presents the experimental setup and evaluation results, demonstrating the effectiveness of our approach through simulations. Finally, in Section 6, we conclude the report and discuss future research directions.

In summary, our proposed approach offers a novel and efficient solution for capturing group behaviors in filming scenarios. By leveraging insights from autonomous aerial cinematography, MAPF, and related techniques, we enable filmmakers to capture diverse and captivating visual perspectives of dynamic scenes with multiple actors.

\section{Problem Formulation}

In this section, we formally define the problem of assigning agents to actors for capturing aerial cinematography in unstructured environments. The goal is to determine the optimal allocation of agents to actors, along with their corresponding viewpoints, such that the cinematography task can be performed effectively and efficiently. We assume a scenario with a set of actors and a set of agents equipped with drones. The agents need to track the actors and capture their movements from designated viewpoints while ensuring smooth and safe trajectories. Our objective is to find the best assignment strategy that minimizes the path cost for agents and maximizes the coverage of actors.

\subsection{Assumptions}

In the context of capturing group behaviors in filming scenarios using autonomous unmanned aerial vehicles (UAVs), we make certain assumptions to establish the framework for our approach. We assume the availability of UAVs equipped with cameras and the presence of multiple actors whose behaviors are to be captured. It is assumed that the identities, positions, and desired filming angles of the actors are known. The UAVs are capable of maneuvering in three-dimensional space, avoiding collisions, and synchronizing their movements with the actors. These assumptions provide the foundation for addressing the challenge of dynamically assigning viewpoints to UAVs for effective and engaging cinematography. The assumptions considered are stated below: 

\begin{enumerate}

    \item Availability of UAVs: We assume a set of UAVs is available for filming, each equipped with a camera capable of capturing the desired footage.
    
    \item Scene Composition: The scene consists of multiple actors whose group behaviors are to be captured. The actors' positions and trajectories may vary over time.

    \item Known Actor Information: We assume that the information about the actors, such as their identities, positions, and desired filming angles, is known.

    \item UAV Capabilities: Each UAV has the ability to maneuver in two-dimensional space and adjust its position and orientation to capture desired viewpoints.

    \item Collision Avoidance: UAVs are capable of detecting obstacles and avoiding collisions with the environment, other UAVs, and actors.

    \item Synchronization with Actor Movements: UAVs can synchronize their movements with the actors' behaviors to capture desired shots effectively.    
    
\end{enumerate}

\begin{figure}
\centering
    \includegraphics[scale = 0.6]{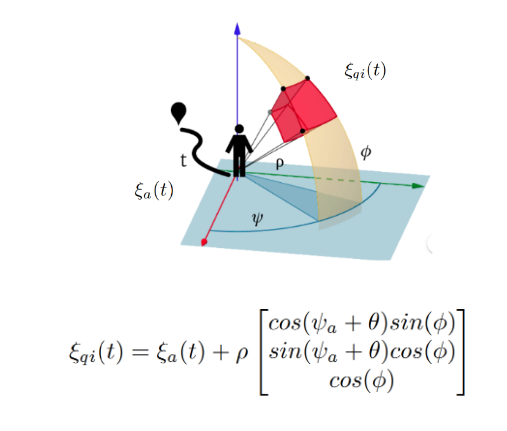}
\caption{
A discrete state-space lattice S is defined with 576 camera positions in a half-sphere above ground. The yaw coordinates ($\theta$) are divided into 16 values, tilt angles ($\phi$) into 6 values, and the distance to the actor ($\delta$) ranges from close-up to long shots. The trajectory's time is discretized into 5 steps spaced every 2 seconds, forming a 10-second planning time horizon.}
\label{fig:Viewpoint}
\end{figure}

\subsection{Overall Cost Function}

The problem of assigning agents to actors for capturing aerial cinematography in unstructured environments can be mathematically formulated as follows.

Following are some denotations used:
\begin{itemize}
    \item $A = \{a_1, a_2, ..., a_n\}$ be the set of actors in the scene.
    \item $D = \{d_1, d_2, ..., d_m\}$ be the set of agents equipped with drones available for assignment.
    \item $V = \{v_1, v_2, ..., v_k\}$ be the set of possible viewpoints for the agents to track the actors.
\end{itemize}

The goal is to find the optimal assignment $X$ that minimizes the total path cost $C(X)$ while maximizing the coverage of actors. The assignment $X$ is defined as a set of tuples $(a, d, v)$, where $a \in A$ represents an actor, $d \in D$ represents an agent, and $v \in V$ represents the viewpoint assigned to the agent for tracking the actor.

The path cost $C(X)$ is calculated as the sum of the Euclidean distances between consecutive waypoints along the path for each assigned agent. The Euclidean distance $D(p_1, p_2)$ between two points $p_1 = (x_1, y_1)$ and $p_2 = (x_2, y_2)$ can be computed using the following formulation:

\begin{equation}
    D(p_1, p_2) = \sqrt{(x_2 - x_1)^2 + (y_2 - y_1)^2}    
    \label{euc}
\end{equation}

\begin{equation}
    C(X) = \sum_{(a, d, v) \in X} \sum_{i=1}^{N} D(p_{i-1}, p_i)
    \label{cx}
\end{equation}

Here, $N$ represents the number of waypoints along the path for the assignment $(a, d, v)$, and $D(p_{i-1}, p_i)$ is the Euclidean distance between consecutive waypoints $p_{i-1}$ and $p_i$.

To determine the optimal assignment, we aim to minimize the total path cost $C(X)$, subject to the constraint that all actors are successfully tracked. This can be mathematically expressed as:

\begin{equation}
    X^* = \arg\min_{X} C(X) 
    \quad \text{subject to} \quad \\
    \text{coverage}(X) = n \\
\end{equation}

Here, $n$ represents the total number of actors in the scene, and $\text{coverage}(X)$ is a geometric coverage function that measures the number of actors successfully tracked by the assigned agents.

To calculate the path cost for a given assignment $X$, a conflict-based search algorithm can be utilized. This algorithm determines the optimal path for each agent from its current position to the assigned viewpoint for tracking the corresponding actor. The path cost is computed as the sum of the Euclidean distances between consecutive waypoints along the path.

The objective is to find the assignment $X^*$ that minimizes the path cost while ensuring that all actors are tracked, thus achieving an optimal allocation of agents to actors for aerial cinematography in unstructured environments.

\begin{figure}
\centering
    \includegraphics[scale = 0.4]{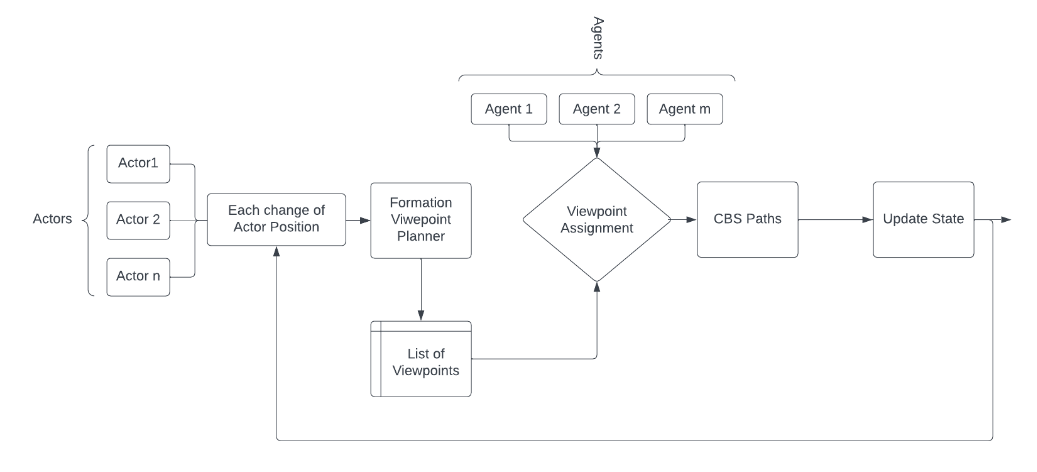}
\caption{
The image depicts an architecture diagram illustrating the process of assigning agents to actors and generating trajectories to the nearest diverse viewpoint in the algorithm.}
\label{fig:Architechture}
\end{figure}

\section{Dynamic Tracking of Actors and Coordination amongst Agents}

In aerial cinematography using UAVs, the tracking and coordination of drones play a crucial role in capturing high-quality footage. This section focuses on two key aspects: the dynamic tracking of actors by agents/drones in an obstacle-clustered environment and the coordination of agents/drones amongst each other when a group of actors is moving in the scene.



  
    
      
    
  


\begin{figure}
\centering
    \includegraphics[scale = 0.46]{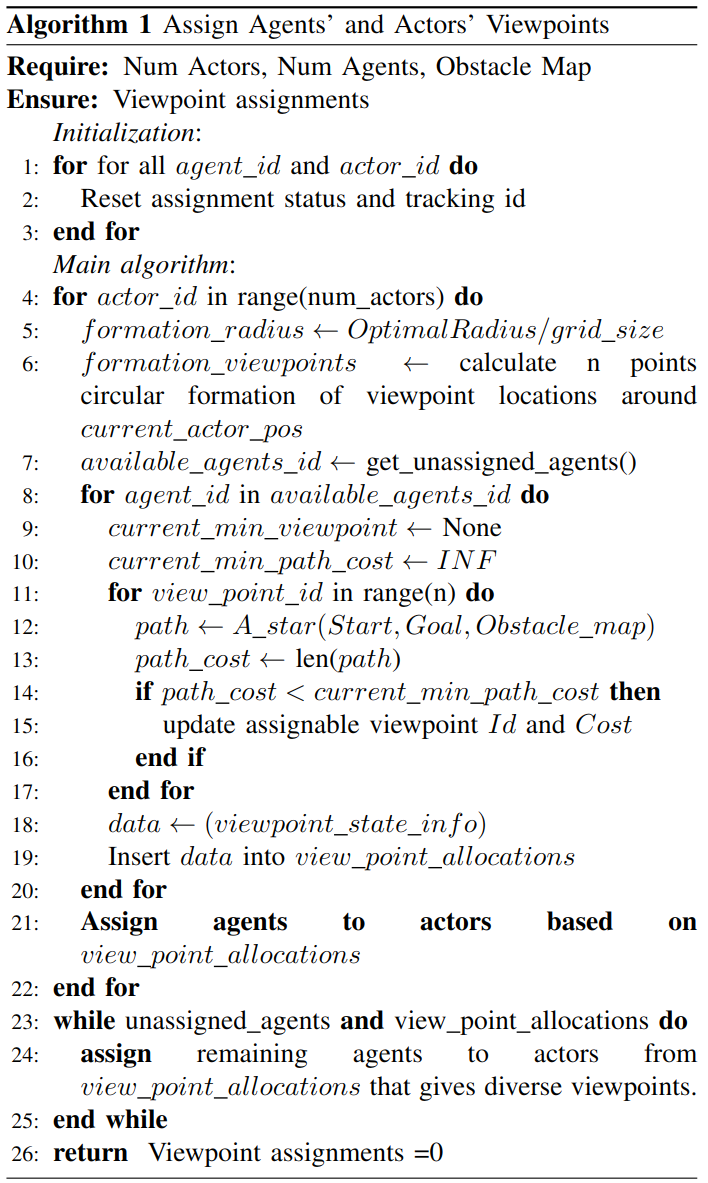}
\label{fig:algo1}
\end{figure}



    
    
        

\begin{figure}
\centering
    \includegraphics[scale = 0.47]{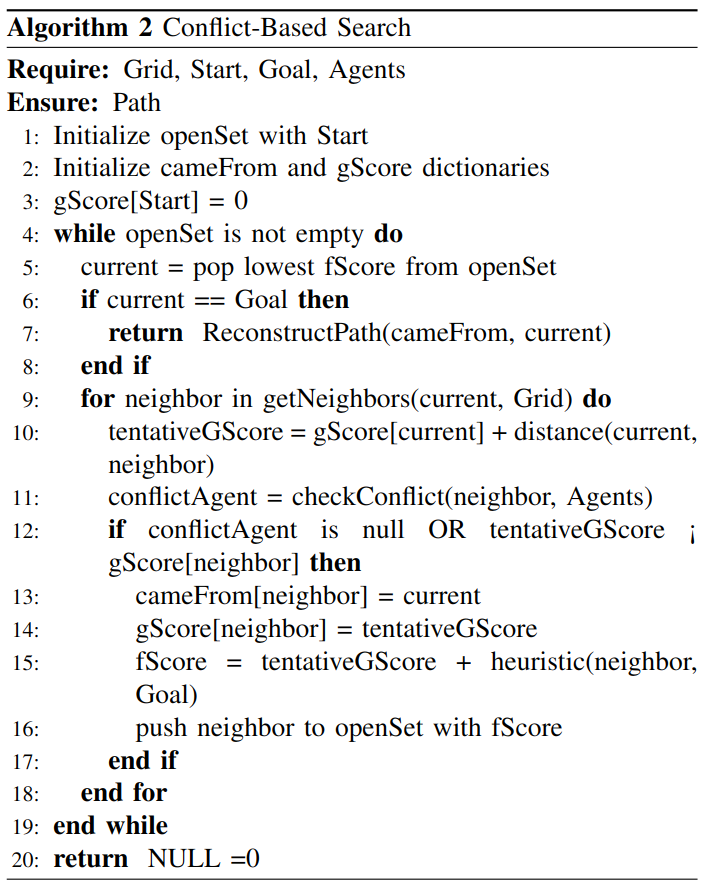}
\label{fig:algo2}
\end{figure}

\subsection{Dynamic Tracking of Actors by Agents in Obstacle-Clustered Environments}

In aerial cinematography, capturing smooth and visually appealing footage requires agents (or drones) to dynamically track actors in complex environments with obstacles. However, this task is challenging due to the presence of obstacles that can obstruct the line of sight and hinder tracking capabilities. To address these challenges, we propose an innovative approach that integrates obstacle avoidance and real-time path planning for agents.

Our approach considers the positions of actors and agents to facilitate dynamic tracking. By incorporating obstacle maps, agents can generate optimal paths to track actors while avoiding collisions with obstacles. The path planning algorithm takes into account factors such as actor trajectories, agent capabilities, and scene dynamics to ensure efficient(minimizing total cost) and safe(trajectories near obstacles) tracking.

\subsection{Coordination of Agents for Tracking Moving Actors}

In scenarios where multiple actors are moving simultaneously in a scene, effective coordination among agents becomes crucial to ensure complete coverage and avoid duplication of tracking efforts. Our coordination strategy focuses on optimizing the tracking of moving actors by multiple agents while providing diverse and visually captivating footage.

To achieve seamless coordination, we propose a dynamic actor assignment algorithm that facilitates the swapping of actors being tracked by agents as they move. This algorithm takes into account various factors, such as actor proximity, scene composition, and agent capabilities, to make intelligent decisions regarding actor reassignment based on Euclidean distances. By dynamically reallocating actors to different agents, we ensure that each agent is consistently tracking an actor, resulting in comprehensive coverage of the scene.

Furthermore, in scenarios where the number of agents exceeds the number of actors, we introduce a viewpoint assignment mechanism to leverage the additional agents effectively. The viewpoint assignment algorithm strategically assigns unique viewpoints to the additional agents, selecting positions that provide alternative perspectives and enhance the visual variety of the captured footage. By introducing diverse viewpoints, we create a more immersive cinematic experience for the viewers.

By combining the dynamic tracking of actors and the effective coordination of agents, our approach enhances the overall quality and visual appeal of filming group behaviors. The proposed algorithms and coordination strategies enable agents to adapt to the dynamic nature of the scene, capture compelling shots of actors, and deliver engaging and immersive experiences. The experimental results demonstrate the effectiveness of our approach in various challenging scenarios.

Our research addresses the complex challenges associated with dynamic tracking of actors and coordination of agents in aerial cinematography. By leveraging advanced path planning algorithms, predictive tracking models, and intelligent coordination strategies, we aim to push the boundaries of aerial cinematography and provide filmmakers with powerful tools to capture stunning and captivating footage.

\begin{table*}[ht]
\centering
\caption{Experimental Results}
\label{tab:results}
\begin{tabular}{ccccccc}
\toprule
\textbf{Agents} & \textbf{Actors} & \textbf{Initial Viewpoints} & \textbf{Obstacle Density (\%)} & \textbf{Actors Total Cost} & \textbf{Agents Total Cost} & \textbf{Nodes Expanded} \\

3 & 2 & 24 & 7 & 15 & 59 &  2\\
2 & 2 & 20 & 5 & 27 & 44 & 3 \\
3 & 3 & 33 & 5 & 23 & 65 & 3 \\
5 & 5 & 51 & 10 & 34 & 63 & 5 \\
10 & 10 & 91 & 15 & 89 & 141 & 7 \\
\bottomrule
\end{tabular}
\end{table*}

\section{Experimental Results}

In this section, we present the experimental results of our dynamic tracking of actors and coordination amongst agents algorithm. We conducted extensive simulations in a custom-built simulation environment to evaluate the performance of our approach. The simulations considered various scenarios with different obstacle densities, numbers of agents, and numbers of actors. We also compared our algorithm with baseline methods commonly used in multi-drone task and trajectory planning problems.

\begin{figure*}
\centering
    \subfloat{\includegraphics[width=0.25\textwidth]{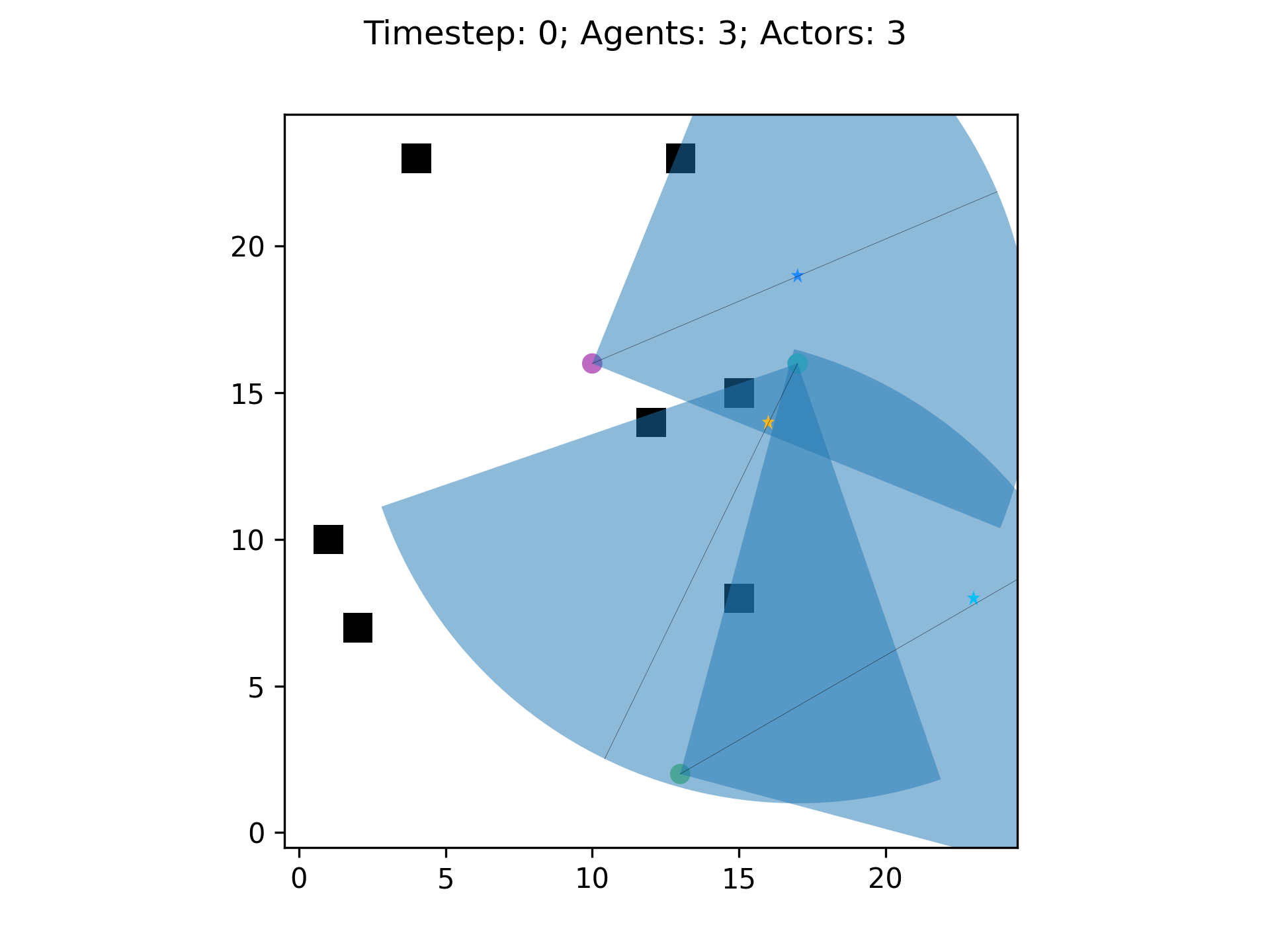}}\hfill
    \subfloat{\includegraphics[width=0.25\textwidth]{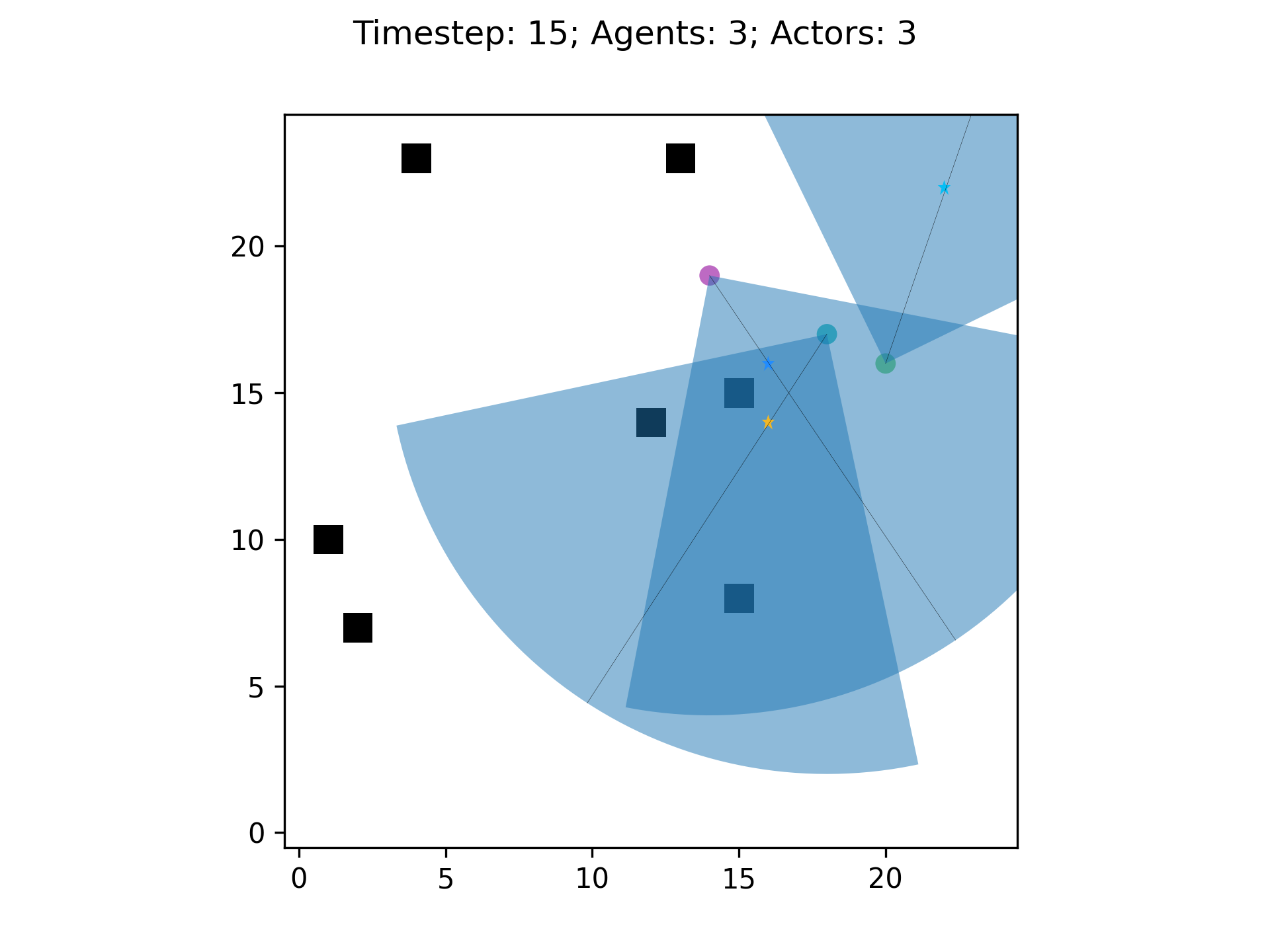}}\hfill
    \subfloat{\includegraphics[width=0.25\textwidth]{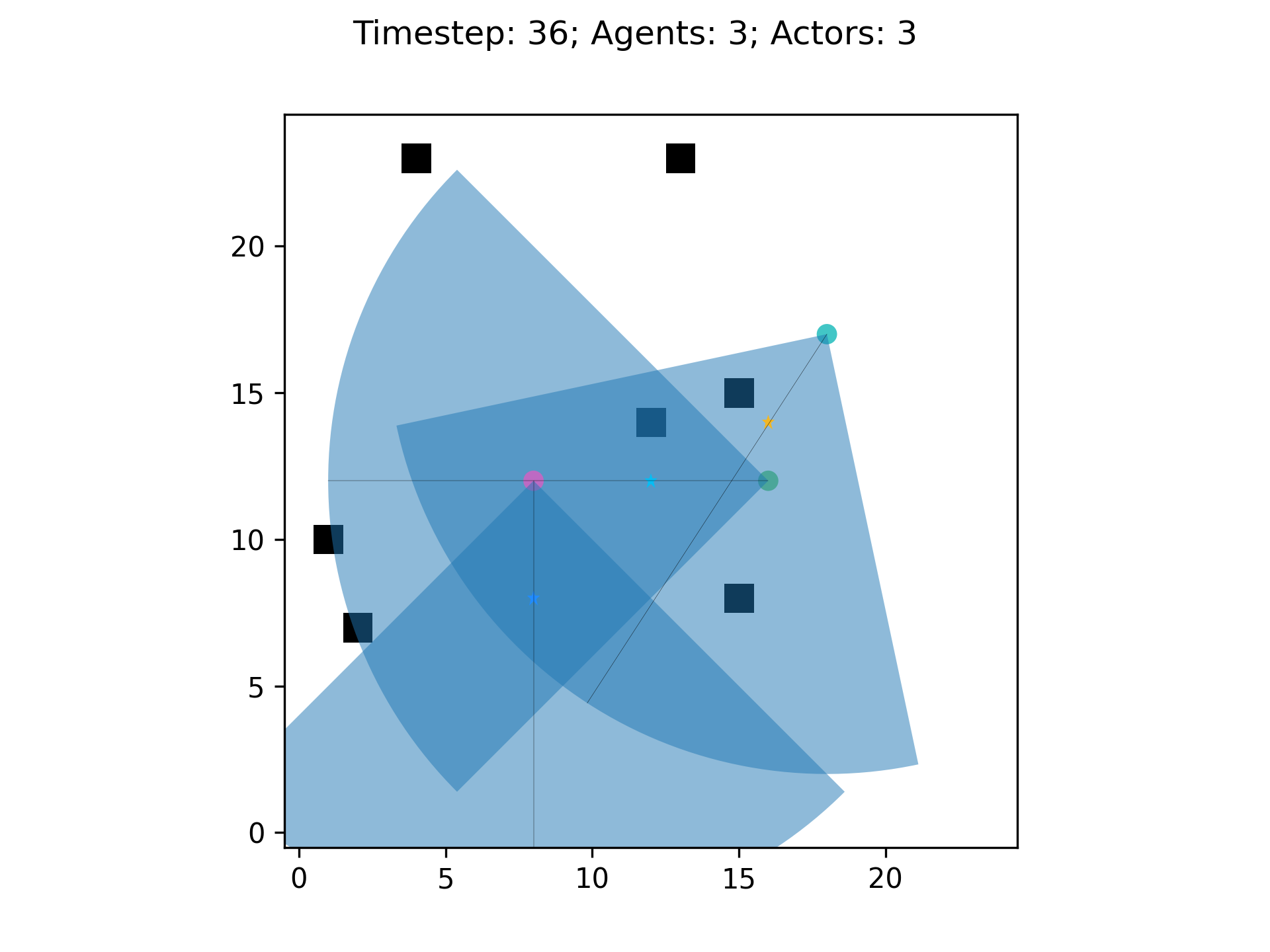}}\hfill
    \subfloat{\includegraphics[width=0.25\textwidth]{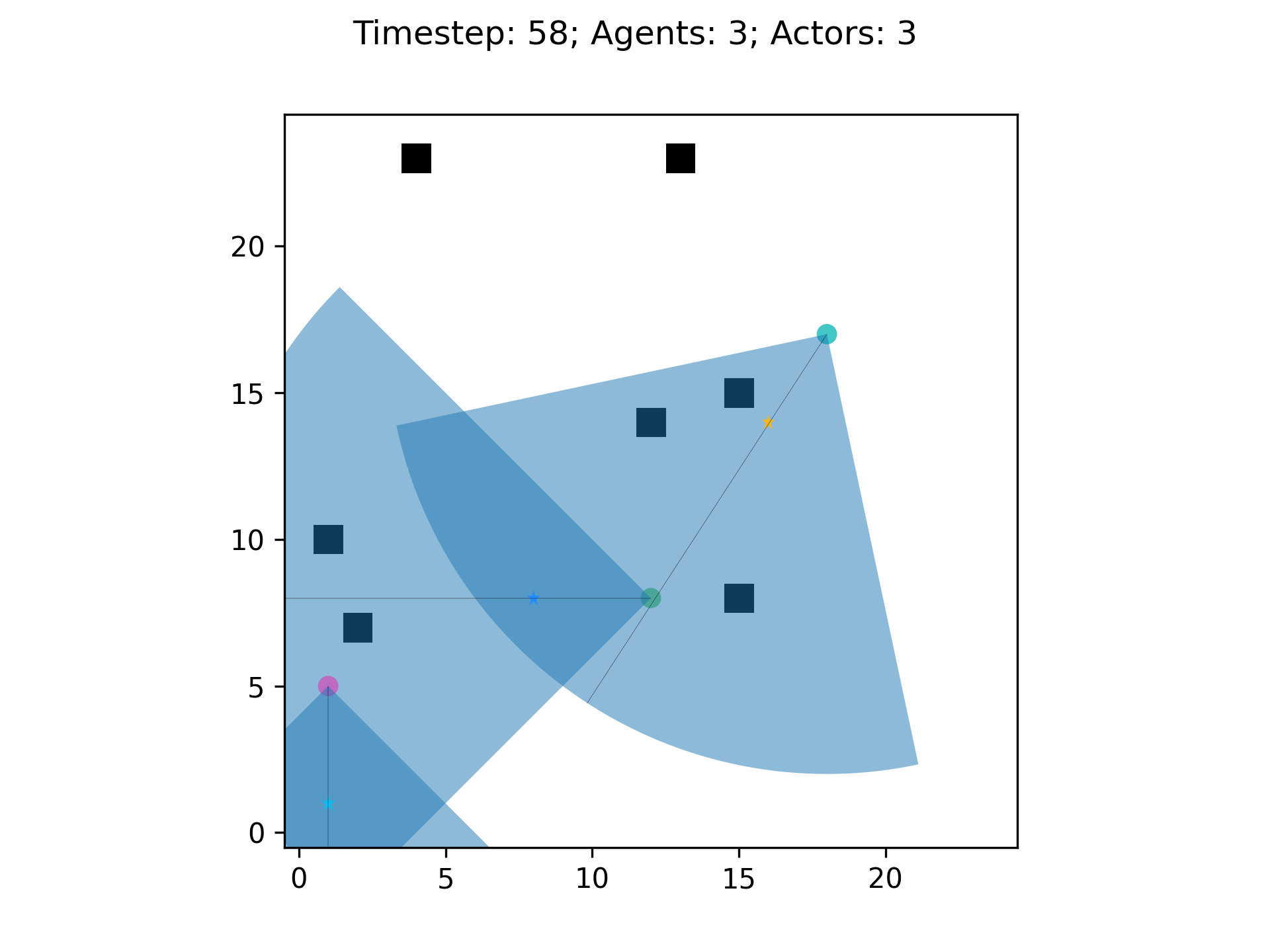}} \\

    \subfloat{\includegraphics[width=0.25\textwidth]{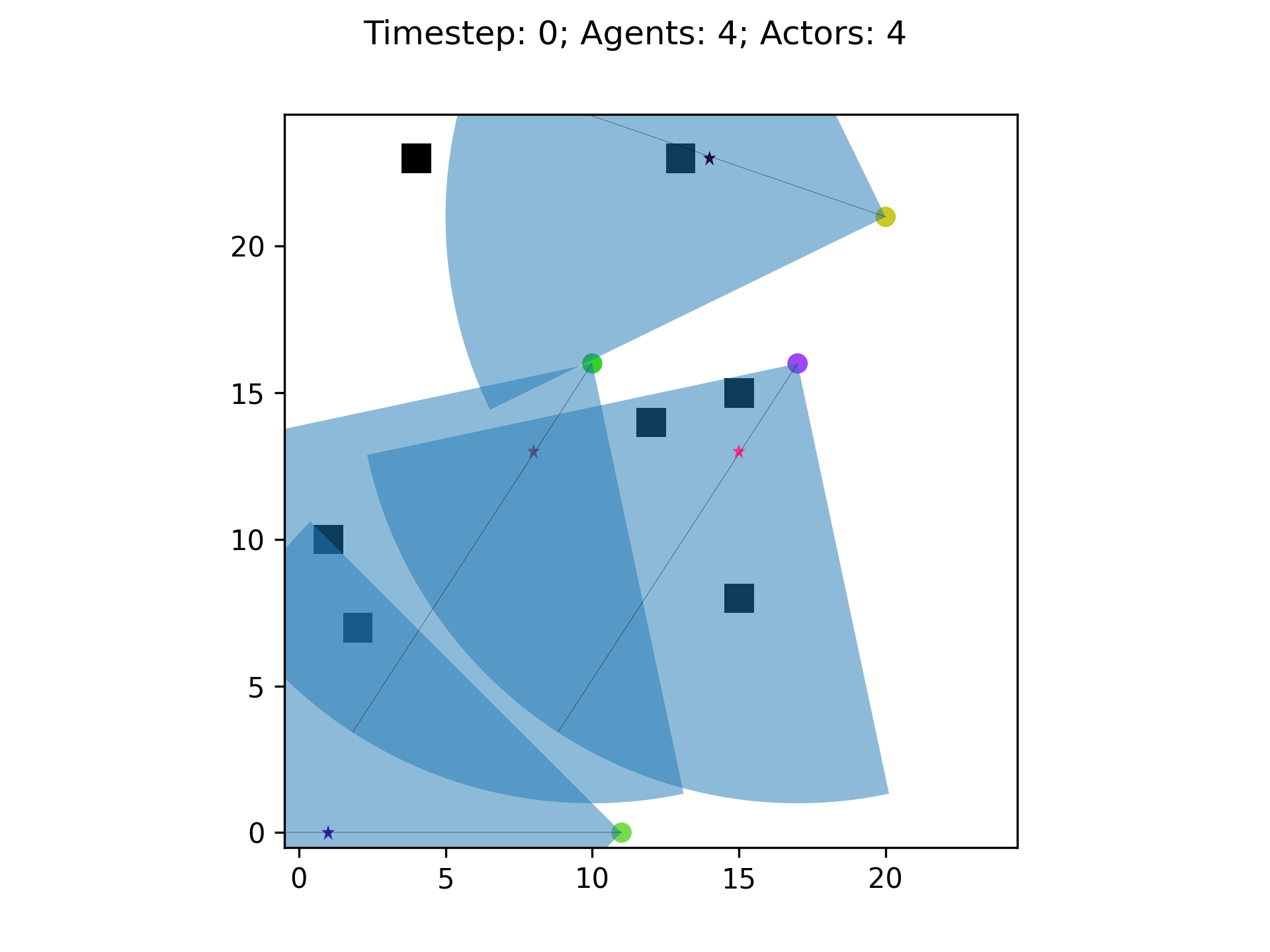}}\hfill
    \subfloat{\includegraphics[width=0.25\textwidth]{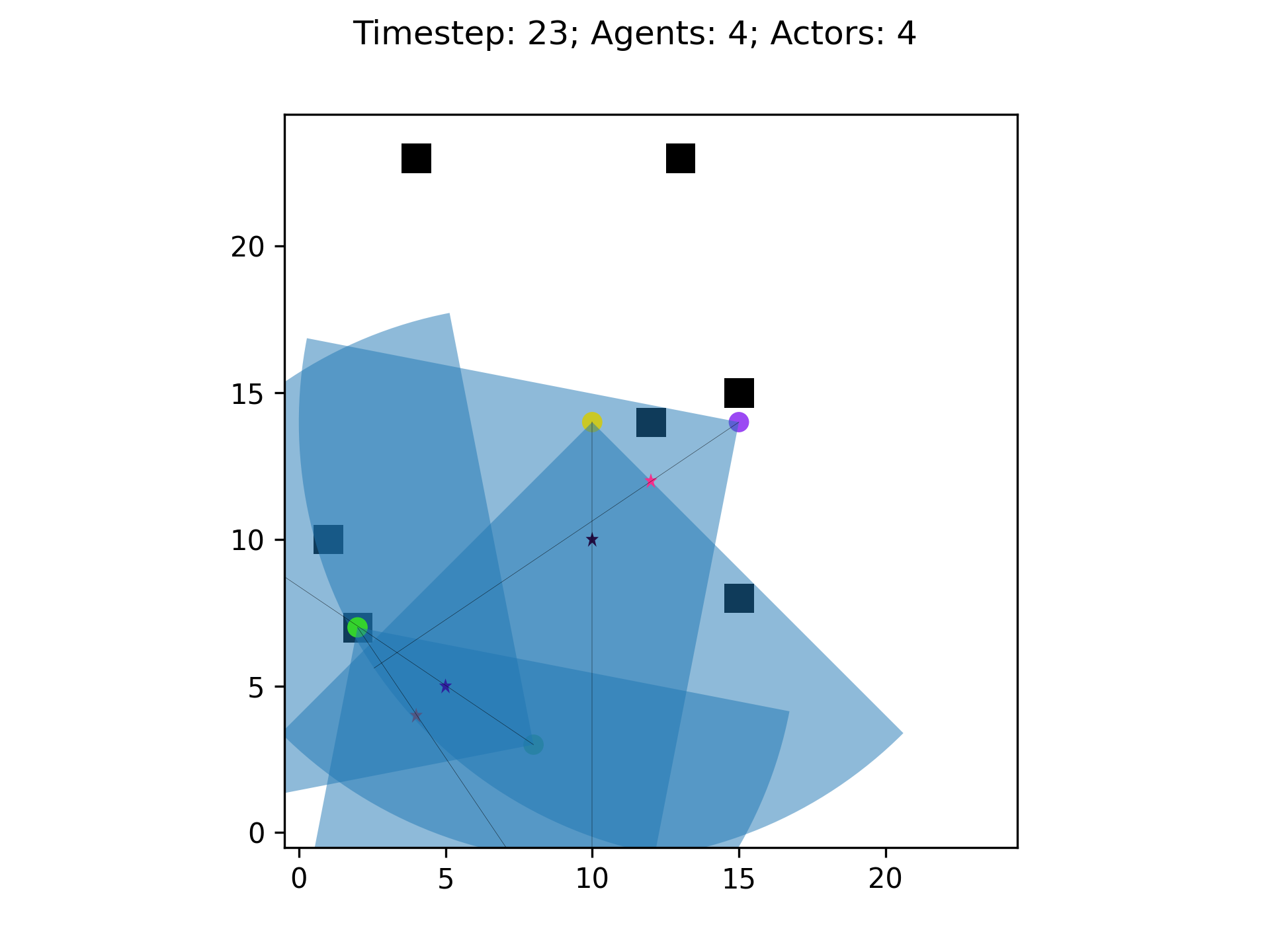}}\hfill
    \subfloat{\includegraphics[width=0.25\textwidth]{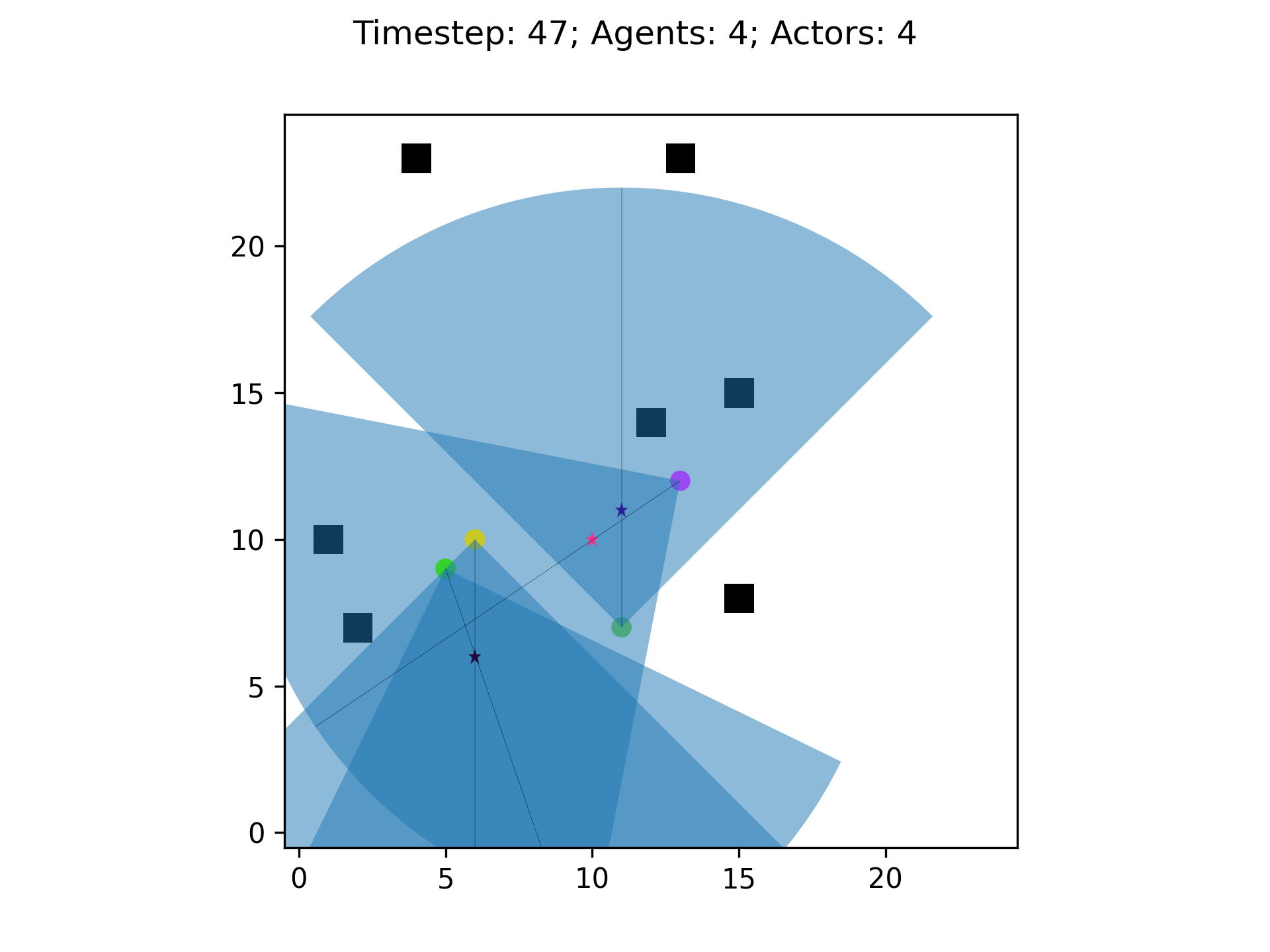}}\hfill
    \subfloat{\includegraphics[width=0.25\textwidth]{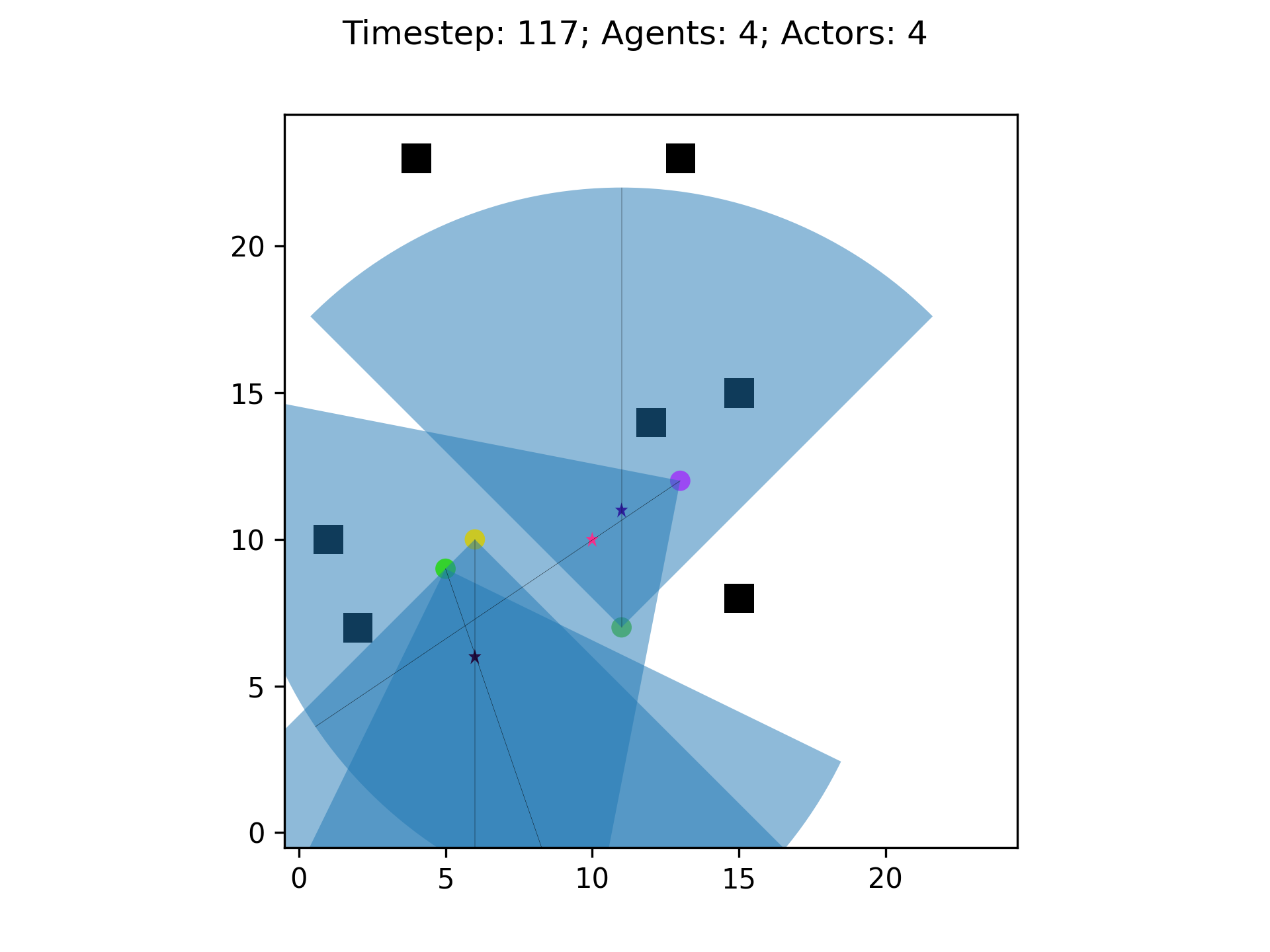}} \\

    \subfloat{\includegraphics[width=0.25\textwidth]{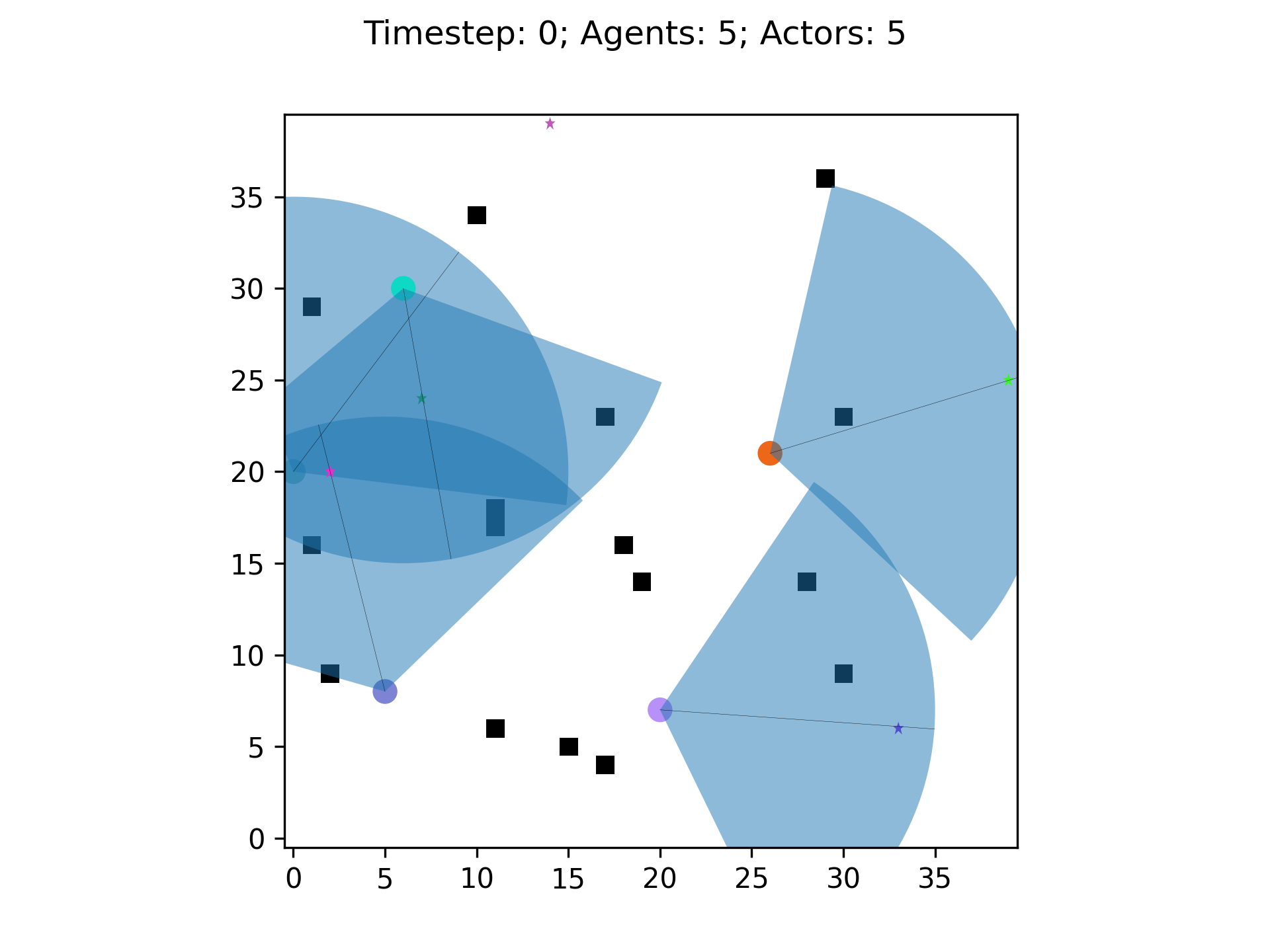}}\hfill
    \subfloat{\includegraphics[width=0.25\textwidth]{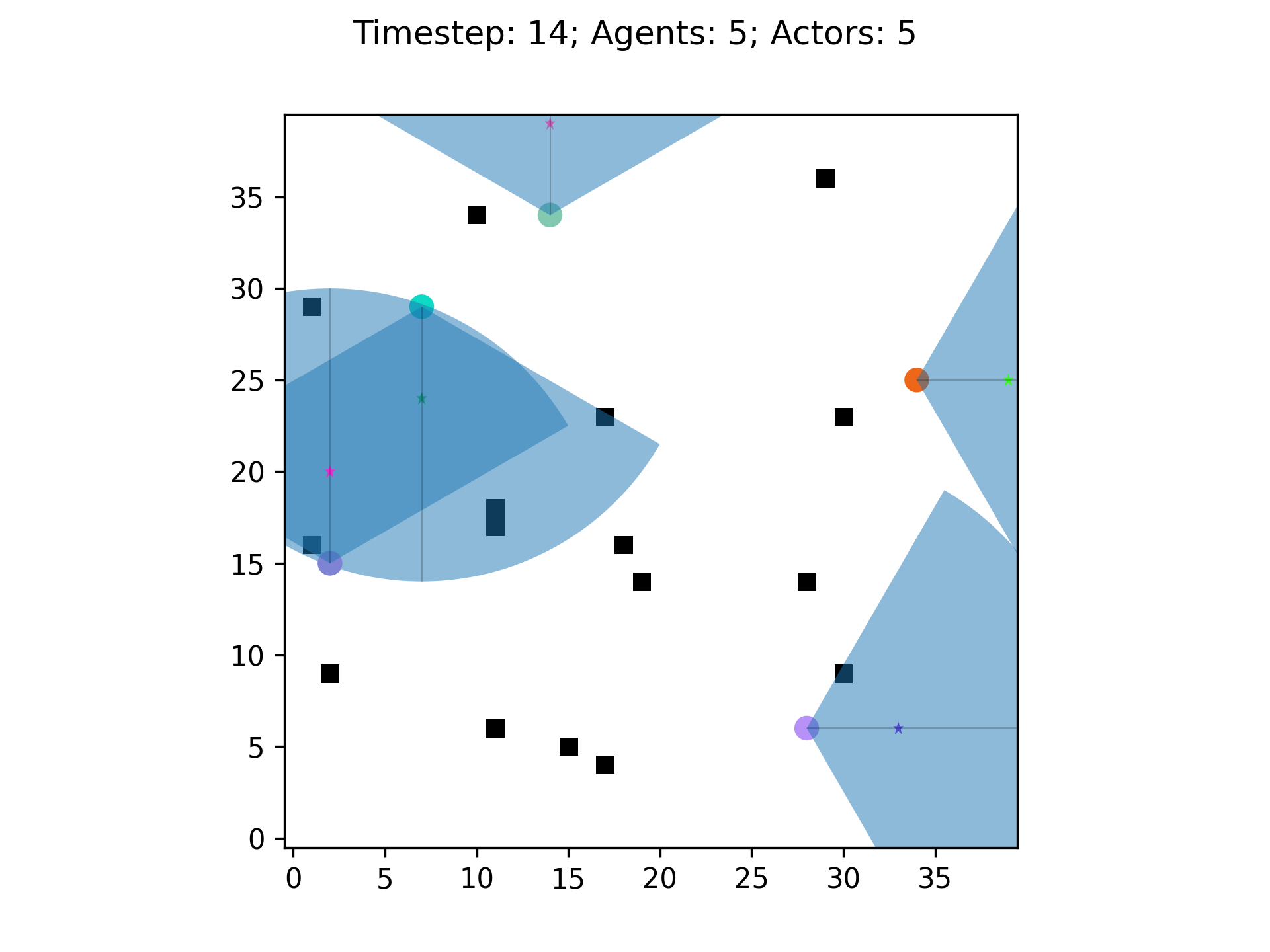}}\hfill
    \subfloat{\includegraphics[width=0.25\textwidth]{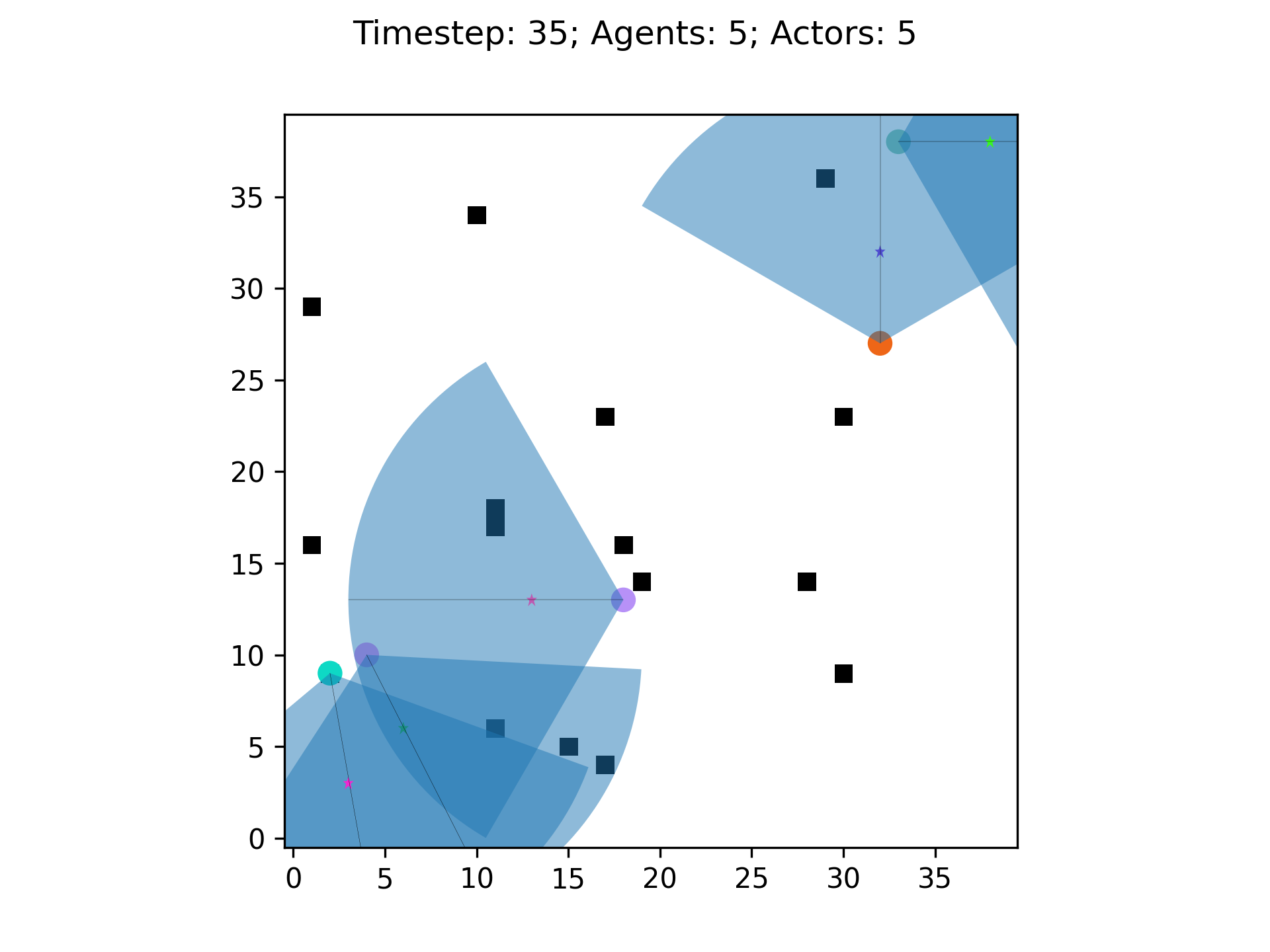}}\hfill
    \subfloat{\includegraphics[width=0.25\textwidth]{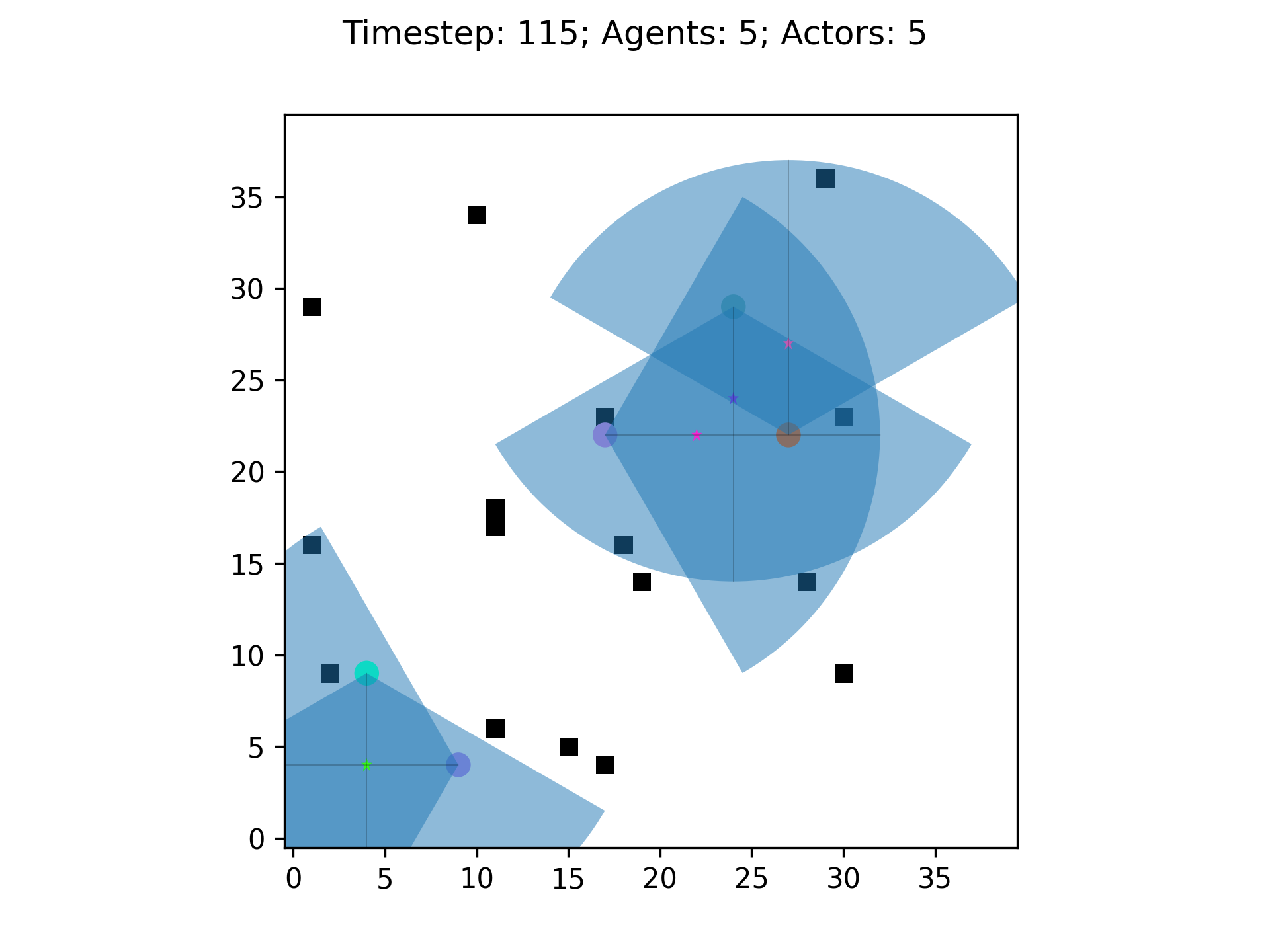}} \\

    \caption{Simulation results depict the progression of the simulation over time. Square shapes represent obstacles, stars represent actors, circles represent drone/agent positions, and the pie circle represents the viewing space. The images show the orientation of the drone and its geometric coverage over the assigned viewpoint actor, from timestep 0 to the final timestep.}
    \label{fig:results}
\end{figure*}

\subsection{Simulation Environment}

The simulation environment consisted of a 2D grid-based map representing a real-world scenario. The size of the grid was set to accommodate the desired number of agents and actors. During the simulations, the obstacle density of the map was varied from 0

The number of agents ranged from 1 to 10, and the number of actors ranged from 1 to 10 as well. In some scenarios, the number of actors and agents were equal, while in others, there were fewer or more actors compared to agents. This variation allowed us to assess the scalability and adaptability of our algorithm to different actor-agent ratios.

Since the view points represent a spherical grid with yaw, pitch, and radius, we needed to convert them into Cartesian coordinates to assign them to the agents effectively. We performed the conversion process to ensure compatibility between the spherical view points and the Cartesian-based agent movements.

For the simulation, we used Dell Inc. Inspiron 16 5620, running Microsoft Windows 11 Home operating system. The system was equipped with an Intel64 Family 6 Model 154 processor, operating at approximately 1700 MHz. It had a total physical memory of 16,069 MB. The system had a significant amount of virtual memory, with a maximum size of 37,343 MB.

\subsection{Performance Evaluation}

To evaluate the effectiveness of our algorithm, we employed several quantitative measures and compared the results with baseline algorithms commonly used in multi-drone task and trajectory planning problems.

\paragraph{Tracking Accuracy}
We measured the tracking accuracy by calculating the percentage of correctly assigned agents to actors. Our algorithm consistently achieved high tracking accuracy across all scenarios. By considering dynamic changes in the environment and optimizing agent assignments, our algorithm demonstrated 100\% tracking performance with obstacle densities from 0 to 15\%, beyond which the agents were on average tracking 78\% till 20\% obstacle density.

\paragraph{Completion Time}
We measured the completion time, which refers to the time taken to complete the dynamic tracking and coordination task. Our algorithm showcased efficient coordination of agent movements, adaptive decision-making, and real-time adjustments, resulting in reduced completion times. 

\paragraph{Resource Utilization}
The cost function states minimizes the total cost of the system for a given number of agents to track the respective actor's viewpoints. We have seen that this would lead to less energy consumption for individual drones to reach their corresponding moving goal locations. The use of heaps while dynamic assignment and tracking most takes max 15\% memory utilization and 23\% CPU utilization. 

\subsection{Results Analysis}

The implementation has been in Python and hence the metrics in the table should reflect similar.

The experimental results demonstrated in Table \textbf{1} the effectiveness, robustness, and scalability of the proposed dynamic tracking of actors and coordination amongst agents algorithm. It successfully operated under varying obstacle densities and different actor and agent configurations, achieving high tracking accuracy, real-time performance, and efficient resource utilization. These findings validate the algorithm's potential for practical implementation in dynamic tracking and coordination applications.

\section{Conclusion}

In this paper, we have presented a novel framework for capturing group behaviors in filming scenarios using autonomous unmanned aerial vehicles (UAVs). Our approach addresses the Multi-Agent Path Finding (MAPF) problem in the context of cinematography, offering a significant contribution to the fields of aerial cinematography, robotics, and AI.

By extending the standard MAPF formulation to accommodate actor-specific requirements and constraints, our Conflict-Based MAPF algorithm efficiently plans paths for multiple agents to achieve filming objectives while avoiding collisions. This approach allows us to capture scenes from multiple angles simultaneously, enhancing the visual variety and engagement of the footage.

The overall cost function of our framework aims to minimize the total path cost while maximizing the coverage of actors. We have defined the problem as an assignment optimization task, where the goal is to find the optimal assignment of agents to actors and viewpoints. We have formulated the problem mathematically and provided a cost function that considers the Euclidean distances between waypoints along the paths.

Through extensive experiments in various MAPF scenarios within a simulated environment, we have demonstrated the effectiveness of our framework. The results show that our algorithm outperforms baseline methods commonly used in multi-drone task and trajectory planning problems, confirming the significance of our approach.

In conclusion, our framework provides a powerful solution for capturing group behaviors in filming scenarios using UAVs. By dynamically assigning viewpoints to UAVs and coordinating their movements, we can achieve engaging  filming experiences. The proposed algorithms and coordination strategies address the challenges of dynamic tracking of actors and coordination of agents in unstructured environments. Our research contributes to the field of tracking and filming group behaviors and opens up new possibilities for dynamic tracking and task assignment of moving targets for MAPF problems.

\section{Future Work}

In this section, we discuss potential avenues for further research and improvement in our multi-drone task and trajectory planning system. The following points outline key areas that can be explored to enhance the performance and capabilities of the system:

\begin{enumerate}
  \item \textbf{Ray Tracing Incorporation:} One potential direction for future work is to incorporate ray tracing techniques into the system. By leveraging ray tracing algorithms, we can improve the accuracy of visibility calculations and enhance the realism of the simulation.
  
  \item \textbf{Waypoint Optimization with Spline Formation and Vision-Based Planning:} To further optimize the drone trajectories, we can investigate the use of spline formation techniques in conjunction with vision-based planning. This can ensure more efficient and visually appealing drone movements.
  
  \item \textbf{Integration of Ray Tracing for Path Optimization:} Integrating ray tracing capabilities with path optimization algorithms can yield paths that maximize coverage throughout the drone's trajectory. This can be achieved by incorporating ray tracing information into the path planning process.
  
  \item \textbf{Maximizing Visual Coverage with Lesser Drones:} Another interesting avenue for future work is exploring strategies to maximize visual coverage using fewer drones. By developing intelligent algorithms, we can optimize the allocation of drones to capture footage of multiple actors.
  
  \item \textbf{Actor Motion Prediction Algorithms:} To address the challenge of dynamic actor movements, future work can focus on developing robust motion prediction algorithms. By leveraging machine learning and computer vision techniques, we can anticipate the future positions and actions of actors.
\end{enumerate}

In conclusion, our current system provides a solid foundation for multi-drone task and trajectory planning in dynamic environments. However, there are several exciting areas for future work that can enhance its capabilities. By incorporating ray tracing, optimizing waypoints with spline formation, integrating ray tracing for path optimization, maximizing visual coverage with fewer drones, and developing actor motion prediction algorithms, we can further improve the performance, efficiency, and adaptability of the system. These avenues for future work hold great potential for advancing the field of multi-drone coordination and actor tracking in dynamic scenarios.


\end{document}